\documentclass[11pt]{article}

\usepackage[preprint]{acl}

\usepackage{times}
\usepackage{latexsym}

\usepackage[T1]{fontenc}

\usepackage[utf8]{inputenc}

\usepackage{microtype}

\usepackage{inconsolata}

\usepackage{graphicx}
\usepackage{booktabs}
\usepackage{multirow}
\usepackage{subcaption}

\usepackage{arydshln}
\newcommand{\mycdashline}[1]{%
    \noalign{\vskip 2pt}  
    \cdashline{#1}        
    \noalign{\vskip 2pt}  
}

\usepackage{amsmath}
\usepackage{amssymb}
\usepackage{bm}

\usepackage{algorithm}
\usepackage[noend]{algpseudocode}

\usepackage[most]{tcolorbox}
\usepackage{listings}
\usepackage{setspace}
\usepackage{color}

\definecolor{isarblue}{HTML}{006699}
\definecolor{isarfaintblue}{rgb}{0.0, 0.75, 1.0}
\definecolor{isargreen}{HTML}{009966}
\definecolor{red}{HTML}{990000}
\definecolor{patriarch}{rgb}{0.5, 0.0, 0.5}

\lstdefinelanguage{isabelle}{%
    keywords=[1]{theory,type_synonym,datatype,fun,abbreviation,definition,proof,lemma,theorem,qed,corollary,have,hence,also,finally,ultimately,moreover,using,\{},
    keywordstyle=[1]\bfseries\color{isarblue},
    keywords=[2]{where,assumes,shows,fixes,and,begin,end,imports},
    keywordstyle=[2]\bfseries\color{isargreen},
    keywords=[3]{if,then,else,case,SOME,let,in,O},
    keywordstyle=[3]\color{isarblue},
    keywords=[4]{ATP},
    keywordstyle=[4]\it\color{patriarch},
    keywords=[5]{show,assume,obtain},
    keywordstyle=[5]\bfseries\color{isarfaintblue},
    keywords=[6]{<proof>},
    keywordstyle=[6]\color{yellow},
}

\lstdefinestyle{isabelle}{%
  language=isabelle,
  escapeinside={&}{&},
  columns=fixed,
  extendedchars,
  basewidth={0.5em,0.45em},
  basicstyle=\singlespacing\ttfamily\small,
  mathescape,
  morecomment=[s][\bfseries\color{red}]{(*}{*)},
  morecomment=[l][\bfseries]{####},
}

\definecolor{mybrown}{RGB}{128,64,0}

\definecolor{codegreen}{rgb}{0,0.6,0}
\definecolor{codegray}{rgb}{0.5,0.5,0.5}
\definecolor{codepurple}{rgb}{0.58,0,0.82}
\definecolor{backcolour}{rgb}{1,1,1}
\lstdefinestyle{mystyle}{
    backgroundcolor=\color{backcolour},
    commentstyle=\color{codegreen},
    keywordstyle=\color{magenta},
    numberstyle=\tiny\color{codegray},
    stringstyle=\color{codepurple},
    basicstyle=\ttfamily\footnotesize,
    breakatwhitespace=false,
    breaklines=true,
    captionpos=b,
    keepspaces=true,
    numbersep=5pt,
    showspaces=false,
    showstringspaces=false,
    showtabs=false,
    tabsize=2
}
\title{
MASA: LLM-Driven Multi-Agent Systems for Autoformalization
}

\author{
  \textbf{Lan Zhang\textsuperscript{1}},
  \textbf{Marco Valentino\textsuperscript{2}},
  \textbf{Andr\'e Freitas\textsuperscript{1,3,4}}\\
  \textsuperscript{1}Department of Computer Science, University of Manchester, United Kingdom\\
  \textsuperscript{2}School of Computer Science, University of Sheffield, United Kingdom\\
  \textsuperscript{3}Idiap Research Institute, Switzerland\\
  \textsuperscript{4}National Biomarker Centre, CRUK Manchester Institute, United Kingdom\\
  \texttt{lan.zhang-6@postgrad.manchester.ac.uk}\\
  \texttt{m.valentino@sheffield.ac.uk}\quad \texttt{andre.freitas@idiap.ch}
}

\begin{document}
\maketitle
\begin{abstract}
Autoformalization serves a crucial role in connecting natural language and formal reasoning. This paper presents MASA, a novel framework for building multi-agent systems for autoformalization driven by Large Language Models (LLMs). MASA leverages collaborative agents to convert natural language statements into their formal representations. The architecture of MASA is designed with a strong emphasis on modularity, flexibility, and extensibility, allowing seamless integration of new agents and tools to adapt to a fast-evolving field. We showcase the effectiveness of MASA through use cases on real-world mathematical definitions and experiments on formal mathematics datasets. This work highlights the potential of multi-agent systems powered by the interaction of LLMs and theorem provers in enhancing the efficiency and reliability of autoformalization, providing valuable insights and support for researchers and practitioners in the field.\footnote{Code and data are available at: \url{https://github.com/lanzhang128/multi_agent_autoformalization}}
\end{abstract}

\section{Introduction}

\begin{figure*}[!t]
    \centering
    \includegraphics[scale=0.41]{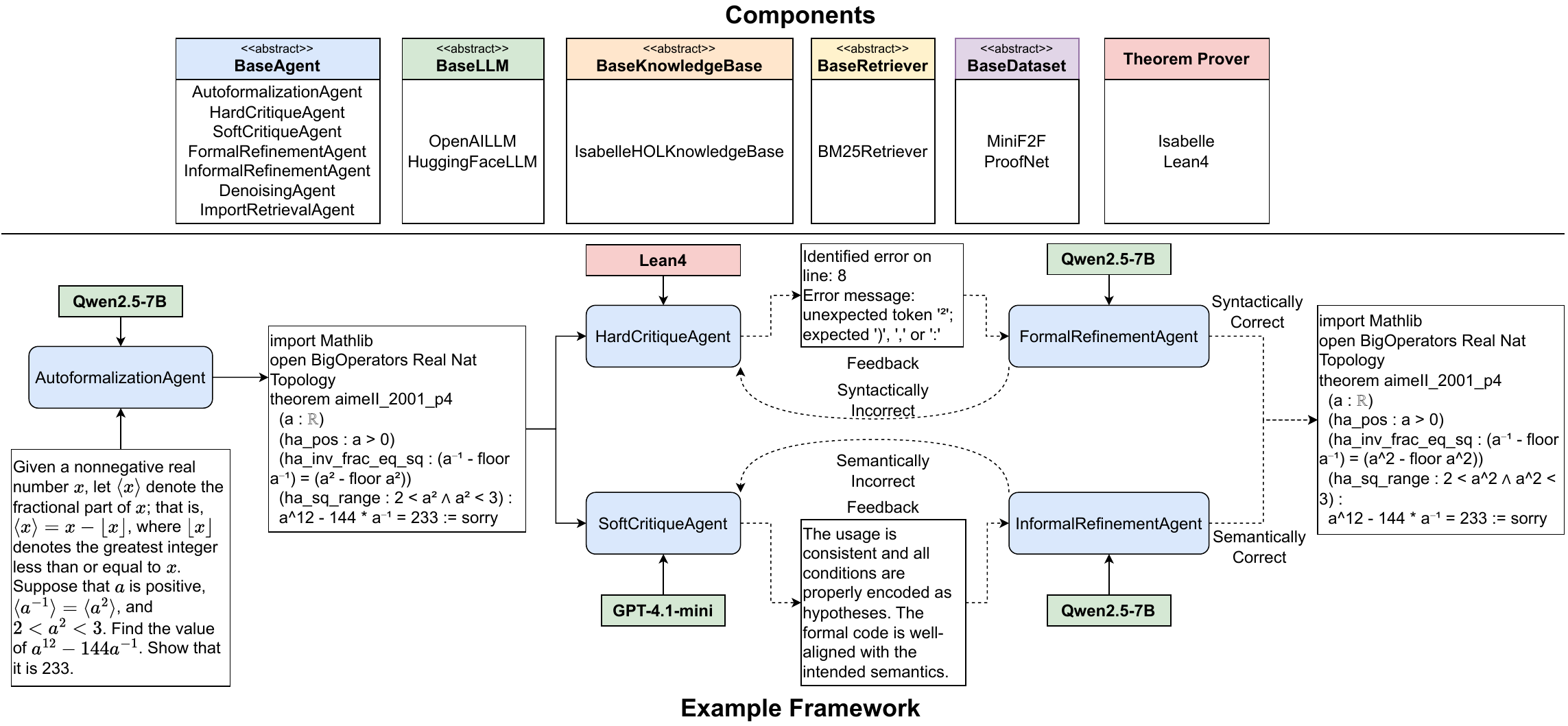}
    \caption{A schematic illustration of MASA. The system comprises agent, LLM, knowledge base, retriever, and theorem prover. The example framework depicts a fully automated pipeline demonstrating how these components interact to formalize a mathematical statement from miniF2F into Lean4.}
    \label{fig:framework}
\end{figure*}

Mathematical reasoning has gained attention in natural language processing~\citep{ferreira2020premise,welleck2021naturalproofs,ferreira-etal-2022-integer,valentino-etal-2022-textgraphs,mishra2022numglue,meadows2023introduction,petersen-etal-2023-neural} and emerged as a core ability of interest in the era of Large Language Models (LLMs)~\citep{wei2022chain,wang2023selfconsistency,meadows-etal-2024-symbolic,lu-etal-2023-survey,mishra-etal-2022-lila}. However, most research on LLM-based mathematical reasoning primarily relies on statements expressed in natural language, resulting in a reasoning process that is neither systematic, transparent, rigorous, nor robust. In contrast, formal mathematics is grounded in a logic-based formal language, where each reasoning step can be systematically verified using theorem provers such as Isabelle~\citep{paulson2000isabelle} or Lean~\citep{lean}.

Although formal mathematics can enhance the systematicity, transparency and rigor of the underlying reasoning process, verifiable mathematical reasoning requires translating natural language representations into formal logical formulas -- a task that demands considerable effort and domain-specific knowledge when done manually. Autoformalization~\citep{wu2022autoformalization}, which aims to automate this process, has shown promising results through prompting with LLMs~\citep{NEURIPS2020_gpt3}. However, addressing real-world autoformalization problems remains a challenging task that can hardly be solved through a single monolithic LLM architecture~\citep{zhang2025autoformalizationwildassessingllms}. This highlights the urgent need to develop a multi-component, distributed system for this task.

Existing implementations of LLM-based autoformalization systems~\citep{wu2022autoformalization, jiang2023draft, zhang-etal-2024-consistent, li2024autoformalize} often lack modularity, flexibility, and extensibility, which hinders researchers from effectively developing and extending systems involving multiple interacting components. In this demonstration paper, we aim to bridge this gap by proposing MASA, a framework that supports the construction of modular agents for building multi-agent systems for autoformalization. Experiments using MASA to build and orchestrate multiple agents demonstrate that the framework is flexible, extensible, and capable of providing valuable insights into multi-agent autoformalization processes.

Our contributions can be summarized as follows:
\begin{enumerate}
    \item We introduce a modular framework for building multi-agent systems for autoformalization, offering flexibility and extensibility for system development.
    \item We showcase the formalization of real-world mathematics using agents, highlighting the practical potential of our framework.
    \item We evaluate our framework in three multi-agent settings, demonstrating its effectiveness and its ability to provide valuable insights into multi-agent autoformalization processes. Our final iterative self-refinement system achieves 35.25\% and 61.89\% formalizations that are both syntactically correct and semantically aligned when applying on Qwen2.5-7B and GPT-4.1-mini, respectively.
\end{enumerate}

\section{Components in MASA}
Autoformalization can be defined as an automatic transformation function $\mathcal{A}$ which maps a natural language statement $s$ to its formal representation $\phi = \mathcal{A}(s)$. A typical approach to this task involves using large language models (LLMs) via prompting~\citep{wu2022autoformalization}, where the function is defined as $\mathcal{A} = \text{LLM}(\text{prompt})$. However, vanilla prompting does not fully exploit various factors that could aid the autoformalization process. A multi-agent setting can incorporate such factors to produce better formalizations. 

To this end, MASA provides support for designing and implementing components that are potentially beneficial for building a multi-agent autoformalization system (Figure~\ref{fig:framework}). We outline each component as follows:

\noindent\textbf{Agent:} The agent is the basic element that performs specific and disentangled functionalities in an interactive manner during the autoformalization process. For autoformalization, the core agents should have the ability to perform tasks such as few-shot autoformalization, providing critiques on formal code, and refining the formalization based on those critiques. Agents with these specialised capabilities can be implemented under an abstract "BaseAgent" class, as illustrated in Figure~\ref{fig:framework}.

\noindent\textbf{Large Language Model (LLM):} LLMs play a key role in an autoformalization system, providing reasoning and linguistic capabilities for implementing specialized agents. Through prompting, LLMs can translate natural language into machine-verifiable languages, provide feedback on formalizations, refine formal codes, etc. The LLM component is abstracted as "BaseLLM" class with specific implementation of OpenAI models\footnote{\url{https://platform.openai.com/docs/api-reference/chat}} and running local HuggingFace moodels\footnote{\url{https://huggingface.co/docs/transformers/llm_tutorial}} in Figure~\ref{fig:framework}.

\noindent\textbf{Knowledge Base (KB):} Formalizations need to align with existing libraries of the formal language. The knowledge base stores information from those libraries so that relevant knowledge can be retrieved to aid an autoformalization process. In Figure~\ref{fig:framework}, the knowledge base is abstracted as "BaseKnowledgeBase" class with an instance of knowledge base of formal statements and proofs from Isabelle/HOL. We provide an example data from the knowledge base in Appendix~\ref{app:kb}.

\noindent\textbf{Retriever:} The retriever ranks the relevance of a formal or informal mathematical statement to data points from a knowledge base of mathematical libraries and retains only the most relevant ones. It is essential for augmenting LLM-generated formalizations. In Figure~\ref{fig:framework}, it is abstracted as "BaseRetriever" class with an implementation of the BM25\footnote{\url{https://pypi.org/project/rank-bm25/}} method.

\noindent\textbf{Theorem Prover (TP):} The theorem prover focuses on the syntactic correctness and underlying logic of a formalization within the relevant formal language. It provides precise information about detected errors for flawed formalizations. The current implementation supports Isabelle via its dedicated  server\footnote{\url{https://isabelle.in.tum.de/dist/Isabelle2025/doc/system.pdf}} and Lean4\footnote{\url{https://github.com/leanprover/lean4}} through REPL\footnote{\url{https://github.com/leanprover-community/repl}}.

\section{A Use Case for Formalizing Real-World Definition with Multiple Agents}\label{sec:agent}

Our framework provides a suite of built-in agents that can be orchestrated to implement autoformalization workflows. To showcase such functionality, we explicitly investigate a use case that focuses on formalizing the definition of softmax\footnote{\url{https://en.wikipedia.org/wiki/Softmax_function}} using multiple agents from MASA.\footnote{Runnable Python Notebook is available at: \url{https://github.com/lanzhang128/multi_agent_autoformalization/blob/main/example_paper.ipynb}} 

\noindent\textbf{Definition of Softmax Function.} Formally, the standard (unit) softmax function \(\sigma\colon \mathbb{R}^K \to (0, 1)^K\), where \(K \ge 1\), takes a vector \(\mathbf{z} = (z_1, \dotsc, z_K) \in \mathbb{R}^K\) and computes each component of vector \(\sigma(\mathbf{z}) \in (0, 1)^K\) with \(\sigma(\mathbf{z})_i = \frac{e^{z_i}}{\sum_{j=1}^K e^{z_j}}\,.\)

The definition is represented as a string, and we aim to formalize it in formal language Isabelle/HOL\footnote{\url{https://isabelle.in.tum.de/dist/library/HOL/index.html}}: 
\begin{lstlisting}[language=python]
informal = "Definition of Softmax Function:..."
formal_language = "Isabelle/HOL"
\end{lstlisting}
We first instantiate an OpenAI model using GPT-4o as the sole backend LLM for agents in Figure~\ref{fig:framework}:
\begin{lstlisting}[language=python]
from llm import OpenAILLM
gpt = OpenAILLM(api_key="...", model="gpt-4o")
\end{lstlisting}

\subsection{Few-Shot Autoformalization Agent}
The autoformalization agent supports the generation of formalization code via an LLM, given a natural language mathematical statement and optional informal-formal statement pairs as exemplars. We can instantiate an autoformalization agent and perform zero-shot autoformalization:
\begin{lstlisting}[language=python]
from agent import AutoformalizationAgent
agent_auto = AutoformalizationAgent(
    llm=gpt, formal_language=formal_language)
zero_formalization, _ = agent_auto(
    informal_statement=informal)
\end{lstlisting}
to obtain the formalization:
\begin{lstlisting}[basicstyle=\small,language=isabelle]
definition softmax :: "real list ⇒ real list" 
    where "softmax z = 
        let exp_z = map exp z; 
            sum_exp_z = sum_list exp_z 
        in map (λzi. exp zi / sum_exp_z) z"
\end{lstlisting}

\subsection{Critique Agent}
The critique agents provide critiques of formalized codes from specific aspects. They are divided into two categories: (1) \textbf{Hard}: critiques are produced by the relevant theorem prover to assess the syntactic aspects of the formalization; (2) \textbf{Soft}: soft-critique agents follow the concept of LLM-as-a-Judge~\citep{zheng2023judging}. Our implementation supports LLMs in producing a binary value (True or False) along with a detailed explanation of the judgment, given the description of an aspect. 

To check the syntactic correctness of the previous formalization, we instantiate a hard critique agent\footnote{Occasionally, the formalization contains only the main body and cannot be tested without additional formatting and necessary imports. The hard critique agent automatically handles this case by adding the required formatting and the default importing statement.}:
\begin{lstlisting}[language=python]
from agent import HardCritiqueAgent
agent_hard = HardCritiqueAgent(
    formal_language=formal_language)
correctness, error_details = agent_hard(
    formalization=zero_formalization)
\end{lstlisting}
we have the critique that the formalization is incorrect with the following detected error detail:
\begin{lstlisting}[basicstyle=\small,language=isabelle]
Undefined type name: "real" Failed to parse type
\end{lstlisting}
This error indicates that "real" in the formalization does not have a reference. We use a tool agent to try to solve this issue.

\subsection{Tool Agent}
Tool agents are designed to address common issues in the autoformalization process using non-reasoning methods. Our implementation includes two such agents: one for denoising~\citep{zhang-etal-2024-consistent}, and another for import retrieval, which tackles the problem of missing relevant imports in formalizations -- often leading to errors that items are treated as undefined~\citep{zhang2025autoformalizationwildassessingllms}.

For the aforementioned issue in the running case, we can instantiate a tool agent for import retrieval:
\begin{lstlisting}[language=python]
from agent import ImportRetrievalAgent
agent_imports = ImportRetrievalAgent(
    formal_language=formal_language, 
    retriever="bm25")
import_formalization = agent_imports(
    formalization=zero_formalization, top_n=1)
\end{lstlisting}
The modified formal code becomes:
\begin{lstlisting}[basicstyle=\small,language=isabelle]
theory Softmax imports Main "HOL.Complex" begin
definition softmax :: "real list ⇒ real list" 
    where "softmax z = 
        let exp_z = map exp z; 
            sum_exp_z = sum_list exp_z 
        in map (λzi. exp zi / sum_exp_z) z"
end
\end{lstlisting}

Using the hard critique agent to test the modified formal code, we obtain the error information:
\begin{lstlisting}[basicstyle=\small,language=isabelle]
Inner syntax error Failed to parse prop
\end{lstlisting}
This error indicates syntax errors in the formalization. Refinement agents are thereby required to improve the formal codes.

\subsection{Refinement Agent}
Refinement agents refine the formalization based on given feedback. Depending on the source of the feedback, they are divided into two categories: (1) \textbf{Formal}: the agent refines the formal codes from a syntactic perspective, given the hard critiques from the relevant theorem prover; (2) \textbf{Informal}: the agent refines the formal codes based on the soft critiques from an LLM.

In the previous case, the formalization has syntax errors. we instantiate a formal refinement agent to let GPT-4o try to fix the errors:
\begin{lstlisting}[language=python]
from agent import FormalRefinementAgent
agent_formal = FormalRefinementAgent(
    llm=gpt, formal_language=formal_language)
formal_refinement, _ = agent_formal(
    informal_statement=informal,
    formalization=import_formalization,
    correctness=correctness,
    error_details=error_details)
\end{lstlisting}
The refined formalization is:
\begin{lstlisting}[basicstyle=\small,language=isabelle]
theory Softmax imports "HOL.Real" begin
definition softmax :: "real list ⇒ real list" 
    where "softmax z = 
        let exp_z = map exp z; 
            sum_exp_z = listsum exp_z 
        in map (λzi. exp zi / sum_exp_z) z"
end
\end{lstlisting}
While the inner syntax error is still not fixed, GPT-4o improves the imports of the formal code to a more concrete one that directly include the definition of "real".

An alternative process of refinement could focus on the semantic aspect. For example, we can describe the aspect of interest as the following:
\begin{lstlisting}[language=python]
aspect_description="whether the formalized code involves all mathematical concepts in the natural language statement."
\end{lstlisting}
and instantiate a soft critique agent to focus specifically on this aspect:
\begin{lstlisting}[language=python]
from agent import SoftCritiqueAgent
agent_soft = SoftCritiqueAgent(
    llm=gpt, formal_language=formal_language,
    aspect_description=aspect_description)
aspect_evaluation, _ = agent_soft(
    informal_statement=informal,
    formalization=import_formalization)
\end{lstlisting}
Although GPT-4o judges this aspect of formalization as correct, we can still instantiate an informal refinement agent to refine the formal code based on the evaluation of the relevant aspect:
\begin{lstlisting}[language=python]
from agent import InformalRefinementAgent
agent_informal = InformalRefinementAgent(
    llm=gpt, formal_language=formal_language)
informal_refinement, _ = agent_informal(
    informal_statement=informal,
    formalization=import_formalization,
    aspect_description=aspect_description,
    aspect_evaluation=aspect_evaluation)
\end{lstlisting}
The direct outputs of aspect evaluation and informal refinement are provided in Appendix~\ref{app:result}.

\section{System Evaluation}\label{section:eval}
The core components of MASA include agents for base autoformalization, critique and refinement, large language models (LLMs), and theorem provers. To evaluate our implementation, we construct three practical multi-agent systems that utilize these agents in conjunction with LLMs and theorem provers. As representative LLMs, we use GPT-4.1-mini, a model from the GPT-4 series~\citep{openai2024gpt4}, and open-sourced Deepseek-Math~\citep{shao2024deepseekmath} and Qwen2.5-7B~\citep{qwen2025qwen25technicalreport}. For benchmarking, we use miniF2F~\citep{zheng2022miniff}, which provides ground-truth formalizations in both Isabelle/HOL and Lean4, and ProofNet\citep{azerbayev2023proofnet}, which provides ground-truth Lean4 code. 

A correctly implemented system should produce formalizations that are both syntactically correct and semantically meaningful. For syntactic correctness, we use pass rate as the evaluation metric. For semantic evaluation, there is currently no universally accepted standard. Thus, we adopt BLEU~\citep{papineni-etal-2002-bleu}, ChrF~\citep{popovic-2015-chrf}, and RUBY~\citep{RUBY} as proxy metrics, following the evaluation setup in~\citep{zhang-etal-2024-consistent}.

Soft-critique agents also can serve as evaluators of semantics for autoformalization. We focus on two aspects of interest for the evaluation: (i) \textbf{Alignment Faithfulness (AF)}: Is the formalized code accurately aligned with the intended semantics of natural language statement? (ii) \textbf{Formalization Correctness (FC)}: Is the formalized code alone valid, nature and well-formed? We employ GPT-4.1-mini as the backend LLM in soft-critique agents to assess the percentage of formalizations that satisfy the relevant evaluation aspect.

\begin{algorithm}[!t]
    \small
    \caption{Hard-Critique Formal Refinement}
    \label{alg:hcfr}
    \begin{algorithmic}[1]
        \State{\textbf{Input:} informal statements $S$, (Optional) informal-formal pairs $\{s,\phi\}$}
        \State{Instantiate an LLM $m_1$}
        \State{Instantiate an autoformalization agent $a_\text{auto}$ with $m_1$}
        \State{Instantiate a hard critique agent $a_\text{hard}$}
        \State{Instantiate an LLM $m_2$}
        \State{Instantiate a formal refinement agent $a_\text{formal}$ with $m_2$}
        \For{$s_i \in S$}
            \State{Formalization $\phi_i=a_\text{auto}(s_i,\{s,\phi\})$}
            \State{Correctness, Error details $c,e=a_\text{hard}(\phi_i)$}
            \If{$c$ is False}
                \State{Refinement $\phi_i=a_\text{formal}(s_i,\phi_i,c,e)$}
            \EndIf
        \EndFor
    \end{algorithmic}
\end{algorithm}

\begin{table}[!t]
    \centering
    \tiny
    \begin{tabular}{l l l c c c c}
        \toprule
        AFA & HCA & FRA & BLEU-4 & ChrF & RUBY & Pass\\
        \midrule
        \multicolumn{7}{l}{\textit{miniF2F-Test (Isabelle/HOL)}}\\
        \midrule
        G4M+ZS & - & - & 26.13 & 33.83 & 41.00 & 65.57\\
        \mycdashline{2-7}
        & Isabelle & G4M & 21.78 & 34.35 & 39.94 & 77.05\\
        \midrule
        G4M+FS & - & - & 32.46 & 45.32 & 47.67 & 76.23\\
        \mycdashline{2-7}
        & Isabelle & G4M & 25.64 & 44.28 & 45.82 & 86.48\\
        \midrule
        DSM+FS & - & - & 6.01 & 35.14 & 28.63 & 29.10\\
        \mycdashline{2-7}
        & Isabelle & DSM & 3.45 & 28.54 & 22.22 & 29.10\\
        \mycdashline{2-7}
        & Isabelle & G4M & 6.33 & 37.28 & 29.20 & 36.48\\
        \midrule
        \multicolumn{7}{l}{\textit{ProofNet-Test (Lean4)}}\\
        \midrule
        G4M+ZS & - & - & 13.98 & 35.80 & 37.34 & 3.30\\
        \mycdashline{2-7}
        & Lean4 & G4M & 11.06 & 34.96 & 34.17 & 3.85\\
        \midrule
        G4M+FS & - & - & 21.35 & 44.45 & 43.81 & 12.09\\
        \mycdashline{2-7}
        & Lean4 &G4M & 18.98 & 44.00 & 41.37 & 14.84\\
        \midrule
        DSM+FS & - & - & 6.81 & 39.67 & 32.19 & 8.79\\
        \mycdashline{2-7}
        & Lean4 & DSM & 3.79 & 34.93 & 18.66 & 8.79\\
        \mycdashline{2-7}
        & Lean4 & G4M & 10.08 & 42.82 & 32.24 & 10.99\\
        \bottomrule
    \end{tabular}
    \caption{Formalization results across different settings. (\textbf{AFA}): Autoformalization Agent; (\textbf{HCA}): Hard Critique Agent; (\textbf{FRA}): Formal Refinement Agent; (\textbf{ZS}): Zero-Shot prompting; (\textbf{FS}): Few-Shot prompting (3 exemplars); (\textbf{G4M}): GPT-4.1-mini; (\textbf{DSM}): Deepseek-Math-7B. All numbers are reported in percentages.}
    \label{tab:hcfl}
\end{table}

\subsection{Hard Critique for Formal Refinement}
The hard critique for formal refinement process (Algorithm~\ref{alg:hcfr}) consists of an autoformalization agent, a hard-critique agent, and a formal refinement agent. Experiments are conducted on miniF2F for Isabelle/HOL and on ProofNet for Lean4. The results are presented in Table~\ref{tab:hcfl}. The few-shot autoformalization agent produces more syntactically correct formalizations compared to the zero-shot setting. The BLEU, ChrF, and RUBY scores also improve in the few-shot setting, indicating that the formalizations are semantically closer to the ground truth. When the formal refinement agent is applied, both zero-shot and 3-shot formalizations demonstrate improved syntactic correctness with GPT-4.1-mini. Although some metric scores decrease after refinement, this is expected, as optimizing for syntactic correctness does not always align with the ground-truth formalization. Deepseek-Math exhibits limited capability in performing formal refinement. However, the stronger LLM, GPT-4.1-mini, is still able to refine formalizations generated by Deepseek-Math. The resulting metric scores are significantly lower than those obtained using GPT-4.1-mini in the zero-shot setting, suggesting that the formal refinement agent modifies the input formalization rather than rewriting it entirely.

\begin{algorithm}[!t]
    \small
    \caption{Soft-Critique Informal Refinement}
    \label{alg:scir}
    \begin{algorithmic}[1]
        \State{\textbf{Input:} informal statements $S$, (Optional) informal-formal pairs $\{s,\phi\}$, aspect description $d$}
        \State{Instantiate an LLM $m_1$}
        \State{Instantiate an autoformalization agent $a_\text{auto}$ with $m_1$}
        \State{Instantiate an LLM $m_2$}
        \State{Instantiate a soft critique agent $a_\text{soft}$ with $m_2,d$}
        \State{Instantiate an LLM $m_3$}
        \State{Instantiate an informal refinement agent $a_\text{informal}$ with $m_3$}
        \For{$s_i \in S$}
            \State{Formalization $\phi_i=a_\text{auto}(s_i,\{s,\phi\})$}
            \State{Explanation, Judgment $e,j=a_\text{soft}(s_i,\phi_i)$}
            \If{$j$ is False}
                \State{Refinement $\phi_i=a_\text{informal}(s_i,\phi_i,d,e,j)$}
            \EndIf
        \EndFor
    \end{algorithmic}
\end{algorithm}

\begin{table}[!t]
    \centering
    \small
    \begin{tabular}{l l l c c}
        \toprule
        AFA & SCA & IRA & AF & FC\\
        \midrule
        DSM+FS & - & - & 38.52 & 47.54\\
        \mycdashline{2-5}
        & G4M+AF & DSM & 47.95 & 38.52\\
        \mycdashline{2-5}
        & G4M+AF & G4M & 90.57 & 79.92\\
        \mycdashline{2-5}
        & G4M+FC & DSM & 43.03 & 52.05\\
        \mycdashline{2-5}
        & G4M+FC & G4M & 77.05 & 86.07\\
        \midrule
        Qwen+FS & - & - & 54.51 & 62.70\\
        \mycdashline{2-5}
        & G4M+AF & Qwen & 73.77 & 66.39\\
        \mycdashline{2-5}
        & G4M+AF & G4M & 93.44 & 85.25\\
        \mycdashline{2-5}
        & G4M+FC & Qwen & 69.26 & 79.92\\
        \mycdashline{2-5}
        & G4M+FC & G4M & 82.79 & 90.57\\
        \bottomrule
    \end{tabular}
    \caption{Lean4 formalization results on miniF2F test set. (\textbf{AFA}): Autoformalization Agent; (\textbf{SCA}): Soft-Critique Agent; (\textbf{IRA}): Informal Refinement Agent; (\textbf{FS}): Few-Shot (3 exemplars); (\textbf{AF}): Alignment Faithfulness; (\textbf{FC}): Formalization Correctness; (\textbf{G4M}): GPT-4.1-mini; (\textbf{DSM}): Deepseek-Math-7B; (\textbf{Qwen}): Qwen2.5-7B. Judges for obtaining scores of AF and FC are based on GPT-4.1-mini. All numbers are reported in percentages.} 
    \label{tab:scir}
\end{table}

\subsection{Soft Critique for Informal Refinement}
The soft critique for informal refinement system (Algorithm~\ref{alg:scir}) comprises three components: an autoformalization agent, a soft-critique agent, and an informal refinement agent. The multi-agent system is evaluated on Lean4 using miniF2F. Results are reported in Table~\ref{tab:scir}. Informal refinement agents are capable of improving specific aspects of formalization. For both Deepseek-Math and Qwen2.5-7B, refinement guided by soft-critique feedback targeting a particular aspect often leads to a significant improvement in that aspect. In most cases, focusing on one aspect also yields improvements in both aspects. The only notable exception is with Deepseek-Math when refining for AF, where improvement in AF results in a decline in FC. Employing a stronger model for refinement, such as GPT-4.1-mini, enhances the overall refinement capability while still preserving the emphasis on the targeted aspect.

\subsection{An Example of Building Multi-Agent System for Iterative Self-Refinement}
\begin{figure}[!t]
    \centering
    \begin{subfigure}{0.4\textwidth} 
      \centering
      \includegraphics[width=\textwidth]{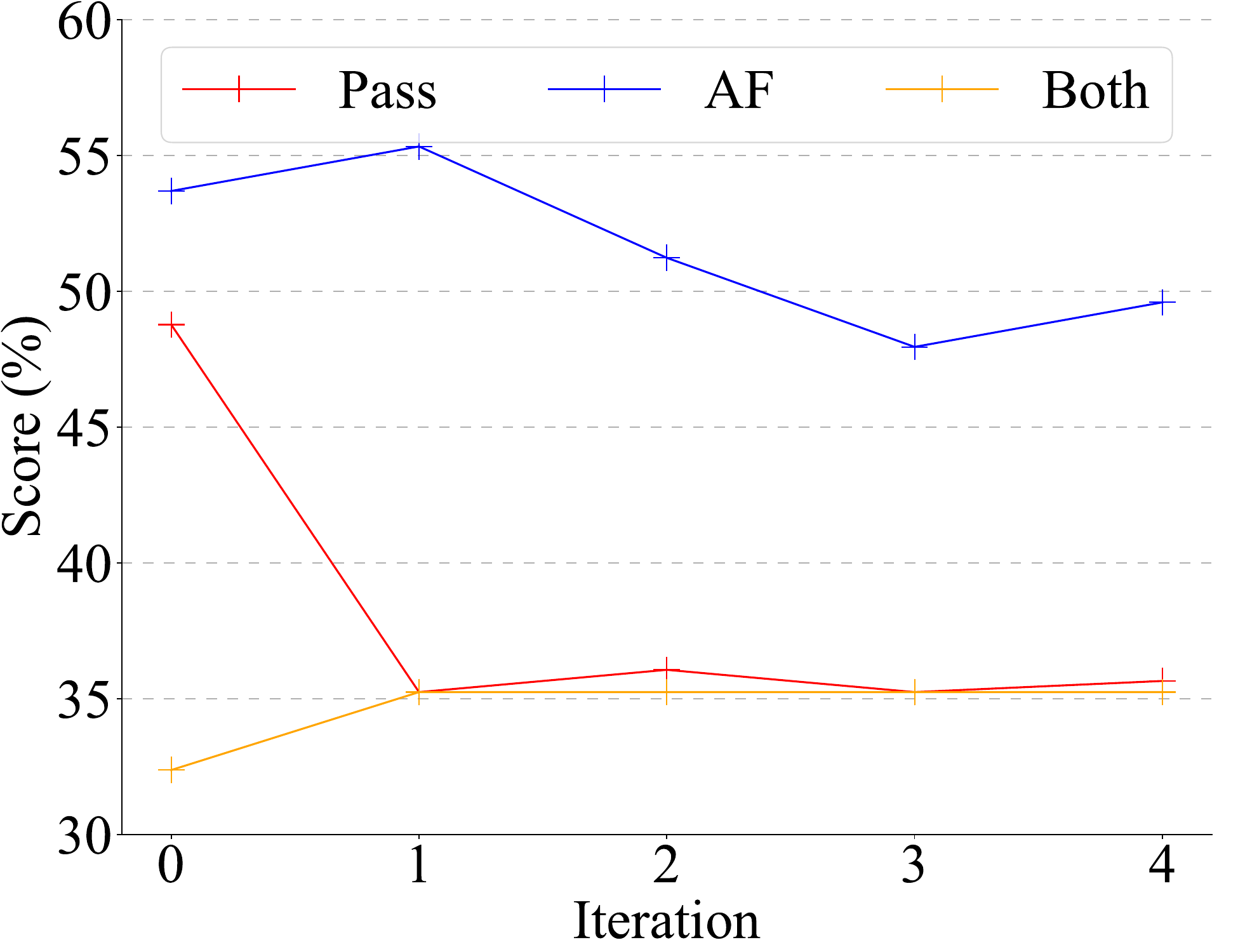}
      \caption{Qwen2.5-7B}
      \label{fig:qwen}
    \end{subfigure}
    \begin{subfigure}{0.4\textwidth} 
      \centering
      \includegraphics[width=\textwidth]{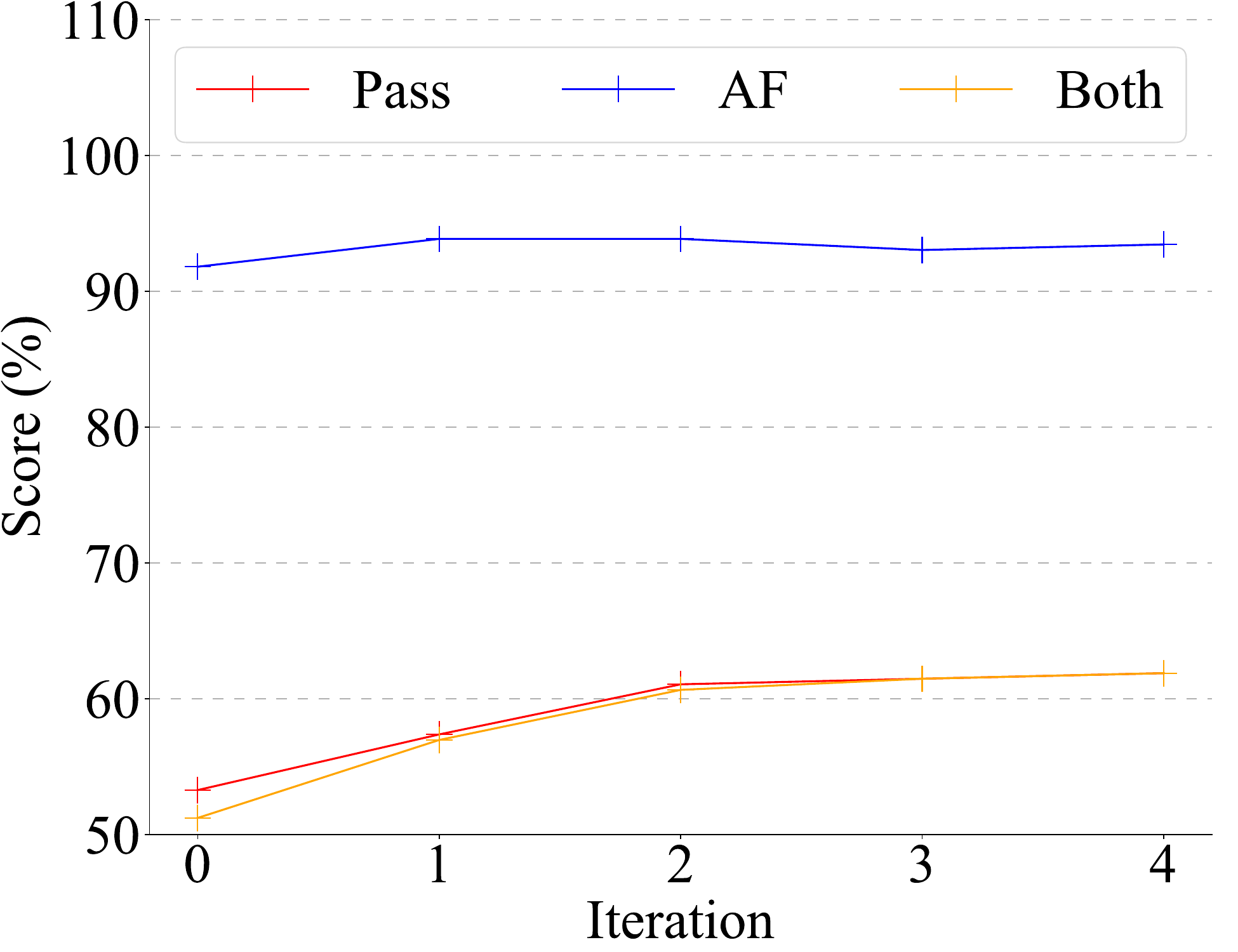}
      \caption{GPT-4.1-mini}
      \label{fig:gpt}
    \end{subfigure}
    \caption{Lean4 formalization results on miniF2F test set with iterative self-refinement.}
    \label{fig:nisr}
\end{figure}

Finally, we present an example of a multi-agent system (Algorithm~\ref{alg:nisr}) that leverages an autoformalization agent, a hard-critique agent, a soft-critique agent, a formal refinement agent, and an informal refinement agent to iteratively improve autoformalization performance using a same backend LLM. We evaluate the system on Lean4 using the miniF2F benchmark. Experiments are conducted with Qwen2.5-7B and GPT-4.1-mini, and we report the pass rate, AF assessment, and the percentage of formalizations that are both syntactically correct and semantically aligned in Figure~\ref{fig:nisr}. We observe that for a smaller model such as Qwen2.5-7B, improvements in syntactic correctness and semantic alignment are more unstable and fluctuate across iterations. In contrast, a stronger model such as GPT-4.1-mini exhibits smoother and more consistent improvements across iterations. Notably, Qwen2.5-7B fails to achieve further gains beyond the first iteration, primarily due to its limited ability to perform formal refinement. On the other hand, GPT-4.1-mini shows consistent improvement with each iteration and is increasingly capable of producing formalizations that are both syntactically correct and semantically aligned.

\begin{algorithm}[!t]
    \small
    \caption{Iterative Self-Refinement}
    \label{alg:nisr}
    \begin{algorithmic}[1]
        \State{\textbf{Input:} informal statements $S$, (Optional) informal-formal pairs $\{s,\phi\}$}, aspect description $d$, number of iterations $n$
        \State{Instantiate an LLM $m_1$}
        \State{Instantiate an autoformalization agent $a_\text{auto}$ with $m_1$}
        \State{Instantiate a hard critique agent $a_\text{hard}$}
        \State{Instantiate an LLM $m_2$}
        \State{Instantiate a soft critique agent $a_\text{soft}$ with $m_2,d$}
        \State{Instantiate a formal refinement agent $a_\text{formal}$ with $m_1$}
        \State{Instantiate an informal refinement agent $a_\text{informal}$ with $m_1$}
        \For{$s_i \in S$}
            \State{Formalization $\phi_i=a_\text{auto}(s_i,\{s,\phi\})$}
            \For{$j=1,\dots,n$}
                \State{Correctness, Error details $c,e=a_\text{hard}(\phi_i)$}
                \If{$c$ is False}
                    \State{Refinement $\phi_i=a_\text{formal}(s_i,\phi_i,e)$}
                \Else
                    \State{Explanation, Judgment $e,j=a_\text{soft}(s_i,\phi_i)$}
                    \If{$j$ is False}
                        \State{Refinement $\phi_i=a_\text{informal}(s_i,\phi_i,d,e,j)$}
                    \EndIf
                \EndIf
            \EndFor
        \EndFor
    \end{algorithmic}
\end{algorithm}

\section{Related Work}
\paragraph{Autoformalization} Autoformalization as an application has demonstrated success in natural language tasks~\cite{quan-etal-2024-enhancing,quan-etal-2024-verification} and formal mathematical reasoning~\citep{jiang2023draft,tarrach2024more}. Recently, with the increasing capabilities of large language models (LLMs), prompting-based methods have shown effectiveness in autoformalizing mathematical statements in Isabelle~\citep{wu2022autoformalization, zhang-etal-2024-consistent, li2024autoformalize,zhang2025autoformalizationwildassessingllms} and Lean~\citep{yang2023leandojotheoremprovingretrievalaugmented,lu2024processdrivenautoformalizationlean4,liu2025rethinking,peng2025criticleancriticguidedreinforcementlearning}. In parallel, \citet{jiang2024multilanguage} and \citet{liu2025atlasautoformalizingtheoremslifting} developed data generation pipelines for constructing large-scale parallel corpora of theorem statements. Despite these advancements, there is a lack of flexible implementations to facilitate a quick start in autoformalization research. Our work aims to address this gap.


\paragraph{Multi-Agent} Multi-Agent systems have emerged as a growing trend with the development of LLMs~\citep{guo2024largelanguagemodelbased}. Researchers have designed systems for applications such as operating systems~\citep{mei2024aiosllmagentoperating}, medical education~\citep{wei2024medcomedicaleducationcopilots}, answer verification~\citep{lifshitz2025multiagent}, and various reasoning tasks, including arithmetic and general reasoning~\citep{li2024more}. However, there have been limited attempts to build multi-agent systems in the context of autoformalization. Our implementation aims to facilitate progress in this area.

\section{Conclusion}
In this work, we present a demonstration of building modular agents for multi-agent systems in autoformalization. Our implementation showcases the potential of collaborative agent-based approaches in tackling the challenges of autoformalization. By providing an accessible and flexible framework, we aim to lower the barrier for researchers and developers interested in exploring this emerging field. Through our demonstration, we illustrate how different agents can work together to iteratively refine and improve formalization outputs. We hope that this demonstration serves as a practical resource for researchers seeking to design and implement their own multi-agent autoformalization systems. We believe this work will contribute to the advancement of AI-driven mathematical reasoning. By fostering collaboration and innovation, we believe this work will contribute to the advancement of AI-driven mathematical reasoning and formal verification.

\section{Limitations}
Despite its contributions, this work has several limitations. First, the proposed multi-agent system lacks a central intelligence agent to distribute and control the different agents. More advanced multi-agent systems still need to be developed, and our implementation can serve as a foundation for such efforts. Additionally, the semantic evaluation of formalizations is limited to high-level judges. Evaluations involving judges with more fine-grained criteria as proxies for semantic analysis are required.

\section*{Acknowledgements}
This work was partially funded by the Swiss National Science Foundation (SNSF) projects NeuMath (200021\_204617) and RATIONAL (200021E\_229196).

\bibliography{custom}

\appendix
\section{A Data Example in the Knowledge Base}
\label{app:kb}
\begin{lstlisting}[language=python]
{"type": "definition", 
"text": "",
"statement": "definition fFalse :: bool where\n\"fFalse \\<longleftrightarrow> False\"\n",
"assumes": "",
"proof": "",
"using": [],
"abs_imports": [
    "HOL.Meson",
    "HOL.Hilbert_Choice"
],
"source": "HOL.ATP",
"id": 0}
\end{lstlisting}

\section{Complementary Output Results}
\label{app:result}

The direct output of aspect evaluation in the main paper is: 
\begin{lstlisting}[basicstyle=\small,language=TeX]
Explanation: The natural language statement describes the softmax function, which maps a vector from \(\mathbb{R}^K\) to \((0, 1)^K\) using the formula \(\sigma(\mathbf{z})_i = \frac{e^{z_i}}{\sum_{j=1}^K e^{z_j}}\). The formal code defines a function `softmax` that takes a list of real numbers and returns a list of real numbers. It uses the `map` function to apply the exponential function to each element of the list, calculates the sum of these exponentials, and then maps each element to its softmax value using the given formula. The code correctly captures the essential components of the softmax function: it computes the exponential of each element, sums these exponentials, and divides each exponential by the sum to produce the output list. The code also correctly handles the input and output types, mapping from a list of real numbers to another list of real numbers, which corresponds to the vector transformation described in the natural language statement. Therefore, the formalized code involves all mathematical concepts in the natural language statement.
Judgement: True
\end{lstlisting}
Informal refinement of this evaluation leads to the formalization:
\begin{lstlisting}[basicstyle=\small,language=isabelle]
theory Softmax
  imports
    Main
    "HOL.Complex"
begin

definition softmax :: "real list ⇒ real list" where
  "softmax z = 
    let exp_z = map exp z;
        sum_exp_z = sum_list exp_z
    in map (λzi. exp zi / sum_exp_z) z"

end
\end{lstlisting}

\section{Examples of Denoising Autoformalizations from Mistral}
\label{app:example}

\textbf{Example 1 (mathd\_numbertheory\_457)}: What is the least positive integer $n$ such that $80325$ divides $n!$? Show that it is 17.

\textbf{Autoformalization:}
\begin{lstlisting}[basicstyle=\small,language=isabelle]
theory aime_1995_p5 imports Complex_Mai
begin
theorem aime_1995_p5:
  fixes n :: nat
  assumes h0 : "80325 | n!"
  shows "17 = n"
  sorry
end
% Note: The statement asks for the least positive integer n such that 80325 divides n!, but the statement provided does not prove that 17 is the answer. It only proves that 17 is a number that makes 80325 divide n!. To prove that 17 is the least positive integer with this property, additional work would be required.
\end{lstlisting}
\textbf{Denoising:}
\begin{lstlisting}[basicstyle=\small,language=isabelle]
theory aime_1995_p5 imports Complex_Main
begin
theorem aime_1995_p5:
  fixes n :: nat
  assumes h0 : "80325 | n!"
  shows "17 = n"
sorry
end
\end{lstlisting}

\textbf{Example 2 (induction\_11div10tonmn1ton)}: Show that for any natural number $n$, we have $11\mid 10^n-(-1)^n$.

\textbf{Autoformalization:}
\begin{lstlisting}[basicstyle=\small,language=isabelle]
theory aime_1995_p1 imports Complex_Main
begin
lemma aime_1995_p1:
  fixes n :: nat
  shows "11 | (pow 10 n) - (- pow 1 + n)"
proof (induction n)
  case 0
  show "11 | 1"
  by (simp add: abs_zero)
  case (Suc n)
  show "11 | (pow 10 (Suc n)) - (- pow 1 + Suc n)"
  have h1 : "11 | (pow 10 n) - (- pow 1 + n)" using n by simp
  have h2 : "(pow 10 n) = 10 * (pow 10 n)" by simp
  have h3 : "(- pow 1 + Suc n) = 11 * (- pow 1 + n)" by simp
  have h4 : "11 | 10 * (pow 10 n) - 11 * (- pow 1 + n)" using h1 by (rule mult_distrib)
  have h5 : "11 | (10 * (pow 10 n) - 10 * (pow 10 n)) + (11 * (- pow 1 + n) - 11 * (- pow 1 + n))" using h4 by (rule add_mono)
  have h6 : "11 | (10 * (pow 10 n) - 10 * (pow 10 n))" using h2 by (rule sub_abs)
  have h7 : "11 | 11 * (- pow 1 + n)" using h3 by (rule sub_abs)
  have h8 : "11 | (10 * (pow 10 n) - 11 * (- pow 1 + n))" using h6 h7 by (rule add_trans)
  qed
end
\end{lstlisting}
\textbf{Denoising:}
\begin{lstlisting}[basicstyle=\small,language=isabelle]
theory aime_1995_p1 imports Complex_Main
begin
lemma aime_1995_p1:
  fixes n :: nat
  shows "11 | (pow 10 n) - (- pow 1 + n)"
sorry
end
\end{lstlisting}
\end{document}